\documentclass[10pt,twocolumn,letterpaper]{article}

\usepackage{iccv}
\usepackage{times}
\usepackage{epsfig}
\usepackage{graphicx}
\usepackage{amsmath}
\usepackage{amssymb}
\usepackage{algorithm}
\usepackage{times}
\usepackage{epsfig}
\usepackage{algpseudocode}
\usepackage{url}
\usepackage{listings}
\usepackage{mathrsfs}
\usepackage{array}
\usepackage{subfigure}
\usepackage{amsthm}
\usepackage{multirow}

\usepackage[pagebackref=true,breaklinks=true,letterpaper=true,colorlinks,bookmarks=false]{hyperref}

\iccvfinalcopy 

\def\iccvPaperID{2441} 

\newcommand{\Eq}[1]  {Eqn.\ (\ref{eq:#1})}
\newcommand{\Eqs}[1] {Eqns.\ (\ref{eq:#1})}
\newcommand{\Fig}[1] {Fig.\ \ref{fig:#1}}

\newcommand{\Sec}[1] {Sec.\ \ref{sec:#1}}

\newcommand{\argmin}{\operatornamewithlimits{argmin}}

\usepackage{color}

\graphicspath{{fig/}}

\ificcvfinal\fi
\begin{document}

\title{Convergence Analysis of MAP based Blur Kernel Estimation} 

\author{Sunghyun Cho\\DGIST\\{\tt\small scho@dgist.ac.kr}
\and
Seungyong Lee\\
POSTECH\\
{\tt\small leesy@postech.ac.kr}}

\maketitle

\begin{abstract}

One popular approach for blind deconvolution is to formulate a maximum a posteriori (MAP) problem with sparsity priors on the gradients of the latent image, and then alternatingly estimate the blur kernel and the latent image.
While several successful MAP based methods have been proposed, there has been much controversy and confusion about their convergence, because sparsity priors have been shown to prefer blurry images to sharp natural images.
In this paper, we revisit this problem and provide an analysis on the convergence of MAP based approaches.
We first introduce a slight modification to a conventional joint energy function for blind deconvolution.
The reformulated energy function yields the same alternating estimation process, but more clearly reveals how blind deconvolution works.
We then show the energy function can actually favor the right solution instead of the no-blur solution under certain conditions, which explains the success of previous MAP based approaches.
The reformulated energy function and our conditions for the convergence also provide a way to compare the qualities of different blur kernels, and we demonstrate its applicability to automatic blur kernel size selection, blur kernel estimation using light streaks, and defocus estimation.

%

\end{abstract}

\section{Introduction}
\label{sec:introduction}

Image blur due to camera shakes is an annoying artifact that severely degrades image quality. Image blur is often modeled as:
\begin{eqnarray}
b = k * l + n ,
\end{eqnarray}
where $b$ is an observed blurry image, $k$ is a blur kernel, $l$ is a latent sharp image, $n$ is noise, and $*$ is the convolution operator.
Blind deconvolution is a problem to estimate $l$ and $k$ from a given blurry image $b$,
which is severely ill-posed because the number of unknowns $l$ and $k$ exceeds the number of observed data $b$.

One popular approach to blind deconvolution is to formulate the problem as a maximum a posteriori (MAP) problem with sparsity priors on the gradients of the latent image, and then alternatingly estimate $k$ and $l$~\cite{Chan98,Shan-SIGGRAPH08,Cho-SIGGRAPHAsia09,Cai-CVPR09,Xu-ECCV10,Xu-CVPR13}.
While several successful MAP based methods with sparsity priors have been proposed, there has been much controversy and confusion about its convergence.
Fergus~et al.~\cite{Fergus-SIGGRAPH06}, in their seminal work, reported that they initially tried a MAP based approach but failed, so adopted a variational Bayesian (VB) approach.
Levin et al.~\cite{Levin-CVPR09} claimed that MAP based approaches with sparsity priors cannot converge to the right solution because sparsity priors favor the no-blur solution, i.e., $k=\delta$, where $\delta$ is a dirac delta function, over the correct one.
To resolve this convergence issue, Krishnan et al.~\cite{Krishnan-CVPR11} introduced a normalized sparsity measure, which favors sharp edges over blurry ones.
Xu et al.~\cite{Xu-CVPR13} claimed that MAP based approaches with an unnaturally sparse image representation can converge to the right solution, and presented a blind deconvolution framework based on an $L_0$ norm based image prior. However, it is not clear whether their successful results are due to either the optimization process, the energy function, or some other factors.

This paper provides an analysis on the convergence of MAP based approaches.
Our analysis explicitly shows that the success of MAP based approaches is due to their energy function favoring the right solution over the no-blur one, and even a na\"{i}ve MAP based approach can converge to the right solution under certain conditions.
For the convergence analysis, we take the most direct approach.
We directly compare the energies of different solutions to find out which solution is favored by the energy function.
We also experimentally analyze conditions for convergence with a large collection of images, and show that the conditions are generally consistent among different images.
Our analysis results support the success of MAP based methods based on extremely sparse image representations, such as \cite{Cho-SIGGRAPHAsia09,Xu-CVPR13}.

To this end, we first introduce a simple modification to a typical joint energy function of $l$ and $k$ and derive an energy function of $k$.
Typical joint energy functions used in previous works involve two variables $k$ and $l$, and this makes it difficult to analyze the energy functions because all possible combinations of $k$ and $l$ should be considered.
Our modification alleviates this by removing one variable from the energy function.
In addition, the reformulated function more clearly reveals how MAP based blind deconvolution works.
Despite the reformulated energy function having only one variable, it is still not straightforward to compare the energies of different solutions.
The reformulated function requires to solve a complex nonlinear optimization problem to compute an energy value,
which makes it impossible to compute the true energy, but only possible to compute an approximate value larger than the true energy in general.
However, we show that it is possible to compute the true energy of the no-blur solution with an energy function of a particular form.
Based on this, our experiments show that the approximate energy of the right solution is still lower than the true energy of the no-blur solution as long as certain conditions are satisfied.


The reformulated energy function and the convergence conditions from our analysis also provide a simple and effective metric to compare the qualities of blur kernels.
We demonstrate that it can be used as a universal metric for solving other problems in deblurring, such as automatic blur size estimation, blur kernel estimation using light streaks, and defocus estimation, which have previously been solved using specifically designed metrics for the problems.

Similar attempts besides our work have been made to unveil the secrets of the success of MAP based approaches.
Perrone and Favaro~\cite{Perrone-CVPR14} claimed that the success of previous MAP based approaches is due to their delayed scaling strategy in the iterative kernel estimation process.
Krishnan et al.~\cite{Krishnan-ArXiv2013} claimed that successful MAP based and variational Bayesian approaches share common components, such as sparsity promotion, $L_2$ norm based priors on the blur kernel, convex sub-problems, and multi-scale frameworks.
However, none of these focused on the energy function, which is the most important factor for blind deconvolution process.

The most relevant to ours is the work of Wipf and Zhang~\cite{Wipf-JMLR14}.
They showed that a VB approach with necessary approximations for making its optimization tractable results in an unconventional MAP approach, where noise level, the latent image, and the blur kernel are coupled together.
They also discussed about the difference of VB and MAP approaches and the convergence of MAP based approaches.
While our work is also on the convergence of MAP based approaches, our work has a few important differences from \cite{Wipf-JMLR14}.
First, we provide a thorough analysis with a number of experimental validations while \cite{Wipf-JMLR14} is completely based on mathematical assumptions and do not provide any experimental results.
Second, in our analysis, we address MAP based blind deconvolution from a perspective of energy minimization, and find conditions for an energy function to favor a sharp solution.
Third, our analysis is based on much simpler and more intuitive equations, which provide a simple and practical guideline to design a MAP based blind deconvolution, e.g., a proper and effective range for the weights of prior terms.
Fourth, our reformulated energy function can be readily utilized for other types of blur kernel estimation problems as we show in \Sec{other_applications}.

\section{Related Work}
\label{sec:relatedwork}


We may categorize recent blind deconvolution methods into mainly three categories.
The first category is MAP based approaches, which alternatingly estimate the latent image and the blur kernel maximizing a joint posterior distribution.
Chan and Wong~\cite{Chan98} alternatingly estimated $k$ and $l$ by minimizing a joint energy function based on total variation.
Shan et al.~\cite{Shan-SIGGRAPH08} introduced a prior on image derivatives based on piecewise continuous polynomials and proposed an efficient optimization method.
While these methods are able to estimate a small scale blur kernel, they often converge to the no-blur solution as shown in \cite{Levin-CVPR09}.
Krishnan et al.~\cite{Krishnan-CVPR11} introduced a normalized sparsity measure that can avoid the no-blur solution, but the measure is highly non-linear, so the method requires a relatively long computation time.
More recently, Xu et al.~\cite{Xu-CVPR13} proposed an approximated $L_0$ norm based prior on image gradients, and showed state-of-the-art results.
Pan et al.~\cite{Pan-CVPR2016} proposed a novel prior to promote sparsity of the dark channel instead of image gradients.
However, despite a number of MAP based approaches having been proposed, it is still unclear how and when these methods converge to the right solution.

The second category is VB based methods, which require marginalization over all possible images.
Fergus et al.~\cite{Fergus-SIGGRAPH06} reported that their initial attempt based on a MAP based alternating estimation failed, as the estimation process either converged to the no-blur solution or diverged, and they presented a VB approach in order to overcome such a convergence problem.
Levin et al.~\cite{Levin-CVPR09} claimed that MAP based approaches with sparsity priors are destined to suffer from the convergence problem because sparsity priors favor blurry images over natural sharp ones, and proposed to use a VB approach. Later, they also introduced an efficient approximation to marginalizing over latent images~\cite{Levin-CVPR11}.
Wipf and Zhang~\cite{Wipf-JMLR14} showed that a VB approach can be recast as an unconventional MAP problem with a particular form of prior that conjoins the latent image, blur kernel, and noise level.
They also provided theoretical analysis about the convergence of MAP based approaches as mentioned earlier.
While VB approaches have proven to be able to estimate accurate blur kernels, they often require complex mathematical derivations, and relatively long computation time even for small images.

The third category uses explicit edge detection such as \cite{Cho-SIGGRAPHAsia09,Xu-ECCV10,Sun-ICCP13}.
They used explicit edge detection in a multi-scale iterative framework to effectively estimate a large blur kernel.
Thanks to their explicit edge detection, these methods can avoid the no-blur solution, and achieve state-of-the-art results in a relatively short computation time.
While these methods involve edge detection, they usually predict sparse and sharp gradient maps of the latent image in their alternating estimation processes, and can still be considered as variants of MAP based approaches.

\section{MAP based Blind Deconvolution}
\label{sec:MAP_based_blind_deconvolution}


Many previous blind deconvolution methods try to estimate a latent image $l$ and a blur kernel $k$ by optimizing the following joint energy function of $l$ and $k$:
\begin{equation}
f(k,l) = \| k * l - b \|^2 + \lambda_l\rho_l(l) + \lambda_k\rho_k(k)
\label{eq:joint_energy_function}
\end{equation}
or its variant.
The first term on the right hand side is a data term, and the second and third terms are prior or regularization terms on $l$ and $k$, respectively.
$\lambda_l$ and $\lambda_k$ are the relative strengths for $\rho_l$ and $\rho_k$, respectively.
For $\rho_l$, sparsity priors have been widely used, such as total variation~\cite{Chan98}, natural image statistics~\cite{Shan-SIGGRAPH08}, and $L_0$-norm based priors~\cite{Xu-CVPR13}.
\Eq{joint_energy_function} can be optimized by alternatingly optimizing two sub-problems:
\begin{eqnarray}
f_l(l;k) &=& \| k * l - b \|^2 + \lambda_l\rho_l(l),~~~~~\textrm{and}
\label{eq:l_step} \\
f_k(k;l) &=& \| k * l - b \|^2 + \lambda_k\rho_k(k).
\label{eq:k_step}
\end{eqnarray}

In this paper, for ease of analysis, we consider a variant of \Eq{joint_energy_function}, which is based on image gradients.
We define $l = \{l_x, l_y\}$, where $l_x$ and $l_y$ correspond to horizontal and vertical gradient maps of the latent image, respectively.
We further assume that $l_x$ and $l_y$ are independent of each other as done in \cite{Cho-SIGGRAPHAsia09,Fergus-SIGGRAPH06,Xu-CVPR13}.
$b = \{b_x, b_y\}$ is defined in the same manner.
We then define each term in \Eq{joint_energy_function} as:
\begin{eqnarray}
\|k*l-b\|^2 &=& \|k*l_x-b_x\|^2+\|k*l_y-b_y\|^2,\\
\rho_l(l) &=& \sum_i \left\{\phi(l_{x,i}) + \phi(l_{y,i})\right\},~~~~~\textrm{and}\\
\rho_k(k)&=&\|k\|^2
\end{eqnarray}
where $i$ is the pixel index.
We define $\phi(x)$ as:
\begin{equation}
\phi(x) = \begin{cases}|x|^\alpha, &\mbox{if}~|x|\ge\tau \\ \tau^{\alpha-2}|x|^2, & \mbox{otherwise}\end{cases}
\end{equation}
so that we can analyze the effects of different sparseness of $\rho_l(l)$ on the convergence of blind deconvolution by changing $\alpha$.
We use $\tau=0.01$ in all our experiments.
While it is more effective to use image intensities and gradients together for blind deconvolution~\cite{Xu-CVPR13}, a gradient based energy function makes it possible to compute the exact global optimum of \Eq{l_step} for $k=\delta$, as we will show later, and consequently makes our analysis easier.


It is known that a na\"{i}ve implementation of \Eq{joint_energy_function} often fails to converge to the right solution, but converges to the no-blur solution.
Levin et al.~\cite{Levin-CVPR09} claimed that this is because of the natures of image blur and sparsity priors.
They showed that image blur has two opposite effects.
First, it makes edges blurry, making image gradients less sparse.
Second, it reduces variance of image gradients, making them sparser.
Previous methods using sparsity priors are based on the first effect, assuming that sharp latent images are mostly piecewise constant with a few step edges.
However, natural sharp images usually have large variance of image gradients even in smooth regions, so the second effect is much stronger than the first one.
Therefore, even though sparsity priors prefer sharp edges to blurry ones in the ideal case, they still prefer a blurry image to a sharp one.

\begin{figure}[t]
\centering
\includegraphics[width=0.95\linewidth]{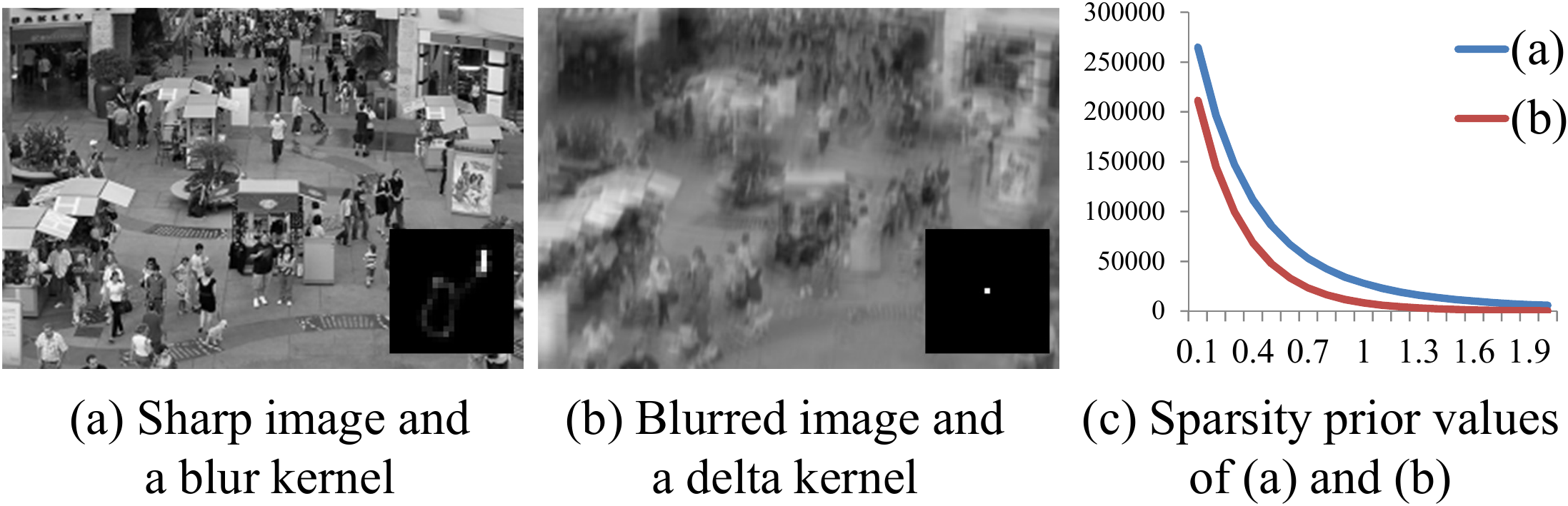}
\caption{The x and y axes of (c) represent different $\alpha$ and sparsity prior values, respectively. While both (a) and (b) produce the exactly same blurred image, the sharp image has higher sparsity prior values for all $\alpha$.}
\label{fig:sparsity_favors_no_blur_solution}
\vspace{-4.7mm}
\end{figure}

\Fig{sparsity_favors_no_blur_solution} describes the aforementioned second effect of image blur.
\Fig{sparsity_favors_no_blur_solution}a is a pair of a sharp image and a blur kernel, which represents a sharp solution, and \Fig{sparsity_favors_no_blur_solution}b is a pair of a blurred image and the delta blur kernel, which represents the no-blur solution.
The sharp solution and the no-blur solution produce the exactly same blurred image.
We then compute their sparsity prior values $\rho_l(l)$ for different $\alpha$.
As described earlier, the sharp solution has higher values for $\rho_l(l)$ compared to the no-blur solution (\Fig{sparsity_favors_no_blur_solution}c), explaining the failure of na\"{i}ve implementations of MAP based approaches.
While this argument seems valid, several works based on MAP based approaches such as \cite{Cho-SIGGRAPHAsia09,Xu-CVPR13} still report good results, which contradict the argument.

\vspace{-0.7mm}
\section{Convergence Analysis}
\label{sec:convergence}

\vspace{-0.7mm}
\subsection{Reformulated Energy Function}
\label{sec:reformulated_energy_function}

In our analysis, to find out which solution the energy function really favors, we take the most direct approach.
We compare the energy values of different solutions.
However, \Eq{joint_energy_function} is not easy to analyze as all possible combinations of $l$ and $k$ need to be considered.
To alleviate this, we first introduce a reformulated energy function derived by embedding \Eq{l_step} into \Eq{joint_energy_function}:
\vspace{-0.7mm}
\begin{eqnarray}
f(k)&=&\min_l f(k,l)
=f(k,\hat{l}_k)\nonumber \\
\vspace{-0.7mm}
 &=& \| k * \hat{l}_k - b \|^2 + \lambda_l\rho_l(\hat{l}_k) + \lambda_k\rho_k(k)
\label{eq:reformulated_energy_function}
\end{eqnarray}
\vspace{-0.7mm}
where
\vspace{-0.7mm}
\begin{eqnarray}
\hat{l}_k=\argmin_l f_l(l;k).
\label{eq:optimal_l_k}
\end{eqnarray}
\vspace{-0.7mm}
\Eq{reformulated_energy_function} is no longer a function of $k$ and $l$, but a function of $k$.
To compute $f(k)$ for a given $k$, we first compute $\hat{l}_k$ in \Eq{optimal_l_k}, and then \Eq{reformulated_energy_function}.
It should also be noted that optimizing \Eq{reformulated_energy_function} is equivalent to optimizing \Eq{joint_energy_function} as we will show in \Sec{global_optimum}.
Consequently, analyzing \Eq{reformulated_energy_function} is equivalent to analyzing \Eq{joint_energy_function}.


Although \Eq{reformulated_energy_function} is now a function of only one variable, it is not feasible to compute the exact energy value of a given $k$ due to the non-convexity of \Eq{optimal_l_k}.
Therefore, in our analysis, we instead compute an approximate energy value.
Specifically, for a given $k$, we first solve \Eq{optimal_l_k} using the iteratively reweighted least squares (IRLS) method~\cite{Levin-SIGGRAPH07}, and obtain an approximate latent image $\hat{l}^\textrm{IRLS}_k$.
Then, we compute an approximate energy $f^\textrm{IRLS}(k)$ by computing \Eq{reformulated_energy_function} with $\hat{l}^\textrm{IRLS}_k$.

\vspace{-3.0mm}
\paragraph{Exact Energy of No-Blur Solution.}
Unfortunately, it is less trustworthy to compare $f^\textrm{IRLS}(k)$ of different $k$ as $f^\textrm{IRLS}(k)$ is only an approximate value, which is always larger than the true energy $f^\textrm{opt}(k)$ for a given $k$.\footnote{Formally, for a given $k$, there exists $\hat{l}^\textrm{opt}=\argmin_l f_l(l;k)=\argmin_l f(k,l)$. By definition, $f(k,\hat{l}^\textrm{opt}) \le f(k,l)$ for all $l$. Consequently, $f^\textrm{IRLS}(k) = f(k,\hat{l}^\textrm{IRLS}) \ge f(k,\hat{l}^\textrm{opt}) = f^\textrm{opt}(k)$.}
Thus, for more accurate analysis, we also compute the exact energy value of the no-blur solution.
Although it is usually impossible to compute the exact energy value of a given $k$ because of the non-convexity of \Eq{optimal_l_k} as mentioned earlier,
as we define our energy function completely based on image gradients, \Eq{optimal_l_k} is pixel-wise independent for $k = \delta$. Therefore, we can find $\hat{l}^\textrm{opt}_\delta$  by solving:
\begin{eqnarray}
\argmin_{l_{*,i} | * \in \{x,y\}} \left| l_{*,i} - b_{*,i} \right|^2 + \lambda_l\phi(l_{*,i})
\label{eq:l_opt_noblur}
\end{eqnarray}
for each pixel of $\hat{l}^\textrm{opt}_{\delta,x}$ and $\hat{l}^\textrm{opt}_{\delta,y}$ independently.
\Eq{l_opt_noblur} can easily be solved using exhaustive search. 

\vspace{-3.0mm}
\paragraph{Analysis}
While \Eq{reformulated_energy_function} is simply a different form of \Eq{joint_energy_function}, \Eq{reformulated_energy_function} more clearly reveals that $\hat{l}_k$ is not an arbitrary natural image, but a {\it sparse} estimate of the latent image $l$ that is coupled with $k$, if $\lambda_l\rho_l(l)$ is strong enough.
In that case, unlike natural sharp images, $\hat{l}$ would have no large variations in smooth regions, but have only flat regions and a few edges.
Then, the sparsity prior term $\rho_l(\hat{l})$ is not affected by the second effect of image blur, but mostly dominated by the first effect.
Consequently, \Eq{reformulated_energy_function} can actually favor a sharp solution over the no-blur one.


To verify this, we compare the energy values of the sharp and no-blur solutions in \Fig{sparsity_favors_no_blur_solution} using the reformulated energy function.
We denote the blur kernels of the sharp solution and the no-blur solution by $k_{gt}$ and $k_\delta$, respectively.
For $k_{gt}$, we first compute the sparse estimate $\hat{l}^\textrm{IRLS}_{gt}$ of the latent image by solving \Eq{optimal_l_k}, and then compute $f^\textrm{IRLS}(k_{gt})$ using \Eq{reformulated_energy_function}.
For $k_\delta$, we compute both approximate and exact latent images ($\hat{l}^\textrm{IRLS}_\delta$, $\hat{l}^\textrm{opt}_\delta$) and their corresponding energy values ($f^\textrm{IRLS}(k_\delta)$, $f^\textrm{opt}(k_\delta)$).

\begin{figure}[t]
\centering
\includegraphics[width=0.95\linewidth]{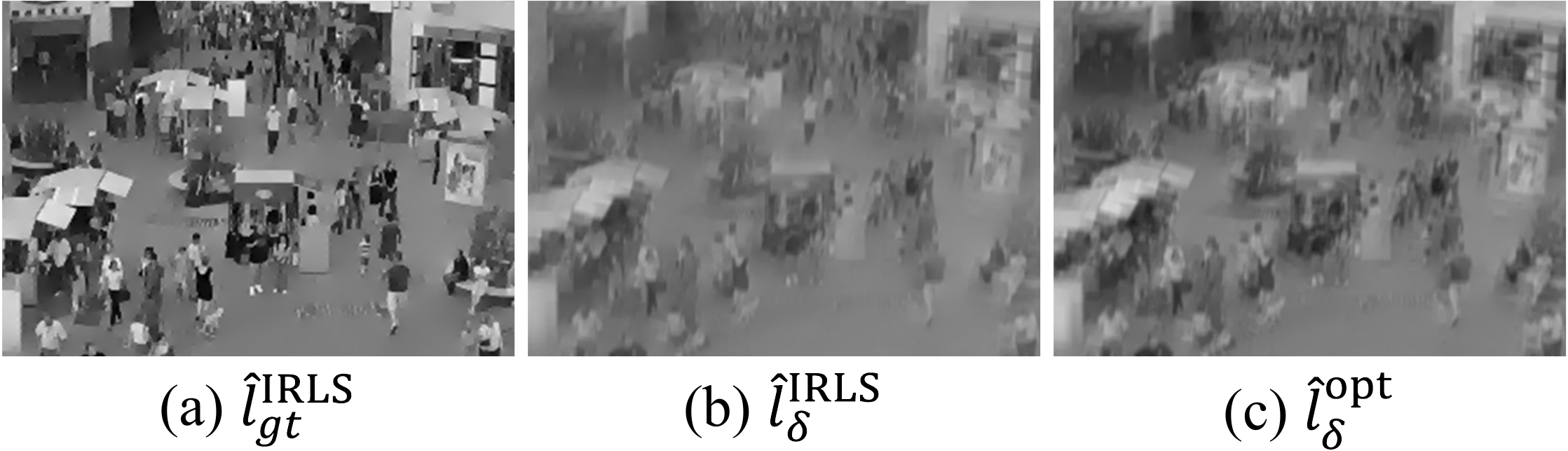}\\
\vspace{5pt}
{\small
\begin{tabular}{|c|c|c|c|}
\hline
& total energy & data & sparsity \\
\hline
$f^{\textrm{IRLS}}(k_{gt})$ & $51.89$ & $31.39$ & $40996.5$  \\
\hline
$f^{\textrm{IRLS}}(k_{\delta})$ & $97.18$ & $55.77$ & $82815.7$  \\
\hline
$f^{\textrm{opt}}(k_{\delta})$ & $74.70$ & $18.67$ & $112053.3$ \\
\hline
\end{tabular}
}
\vspace{5pt}
\caption{Top row: sparse estimates of the latent image for the ground truth kernel and the delta kernel. As our energy function is defined using image gradients, latent image estimates are gradient maps. We visualize them using Poisson image reconstruction, which restores intensities from image gradients, as done in~\cite{Fergus-SIGGRAPH06}. Bottom row: energy values, data terms, and sparsity priors of the ground truth blur kernel $k_{gt}$ and the delta kernel $k_{\delta}$. We set $\alpha = 0.1$ and $\lambda_l = 0.0005$. }
\label{fig:sparse_estimates}
\vspace{-2.5mm}
\end{figure}

\Fig{sparse_estimates} shows the computed latent images and the energy values of $k_{gt}$ and $k_\delta$.
As discussed above, the sparse estimates $\hat{l}_{gt}^\textrm{IRLS}$, $\hat{l}_\delta^\textrm{IRLS}$ and $\hat{l}_\delta^\textrm{opt}$ have only smooth regions and a few edges together with almost no variation in smooth regions.
$f^\textrm{IRLS}(k_{gt})$ and $\rho_l(\hat{l}_{gt}^\textrm{IRLS})$ are also smaller than $f^\textrm{IRLS}(k_{\delta})$ and $\rho_l(\hat{l}_{\delta}^\textrm{IRLS})$, respectively.
More importantly, $f^\textrm{IRLS}(k_{gt})$ and $\rho_l(\hat{l}_{gt}^\textrm{IRLS})$ are smaller than $f^\textrm{opt}(k_{\delta})$ and $\rho_l(\hat{l}_{\delta}^\textrm{opt})$, respectively, even though $\hat{l}_{gt}^\textrm{IRLS}$ is an approximate estimate.
This result means that the global optimum of \Eq{reformulated_energy_function}, which is equivalent to the global optimum of \Eq{joint_energy_function}, favors the sharp solution over the no-blur solution.

\subsection{Conditions for Avoiding No-Blur Solution}
\label{sec:no_blur_condition}

In this subsection, we analyze when MAP based approaches converge to the right solution.
To this end, we consider the following two conditions.
\vspace{-0.7mm}
\begin{eqnarray}
 f(k_{gt}) / f(k_{\delta}) &<& 1,~~~~~\textrm{and}
\label{eq:condition1_for_avoiding_no_blur}  \\
  \rho_l(\hat{l}_{gt}) / \rho_l(\hat{l}_\delta) &<& 1 .
\label{eq:condition2_for_avoiding_no_blur}
\end{eqnarray}
\vspace{-0.7mm}
While the first condition is sufficient for avoiding the no-blur solution, we also consider the second one because the prior $\rho_l$ is the key to distinguish between sharp and blurry latent images.
To satisfy the second condition, the latent image estimates $\hat{l}_{gt}$ and $\hat{l}_\delta$ should be sparse enough as shown in \Sec{reformulated_energy_function}.
This means that $\lambda_l$ should be appropriately large and $\alpha$ should be small.
If $\lambda_l$ is too small, then $\hat{l}_{gt}$ will be similar to a natural sharp image, which is not sparse but has large variation in smooth regions, and the second effect of blur discussed in \Sec{reformulated_energy_function} will kick in.
On the other hand, too large $\lambda_l$ will make $\hat{l}_{gt}$ and $\hat{l}_\delta$ entirely flat images with no edges at all, so they will be indistinguishable.
Larger $\alpha$ will also produce blurrier edges on both $\hat{l}_{gt}$ and $\hat{l}_\delta$, making them less distinguishable.

\begin{figure}[t]
\centering
{\small
\begin{tabular}{cc}
\includegraphics[width=0.43\linewidth]{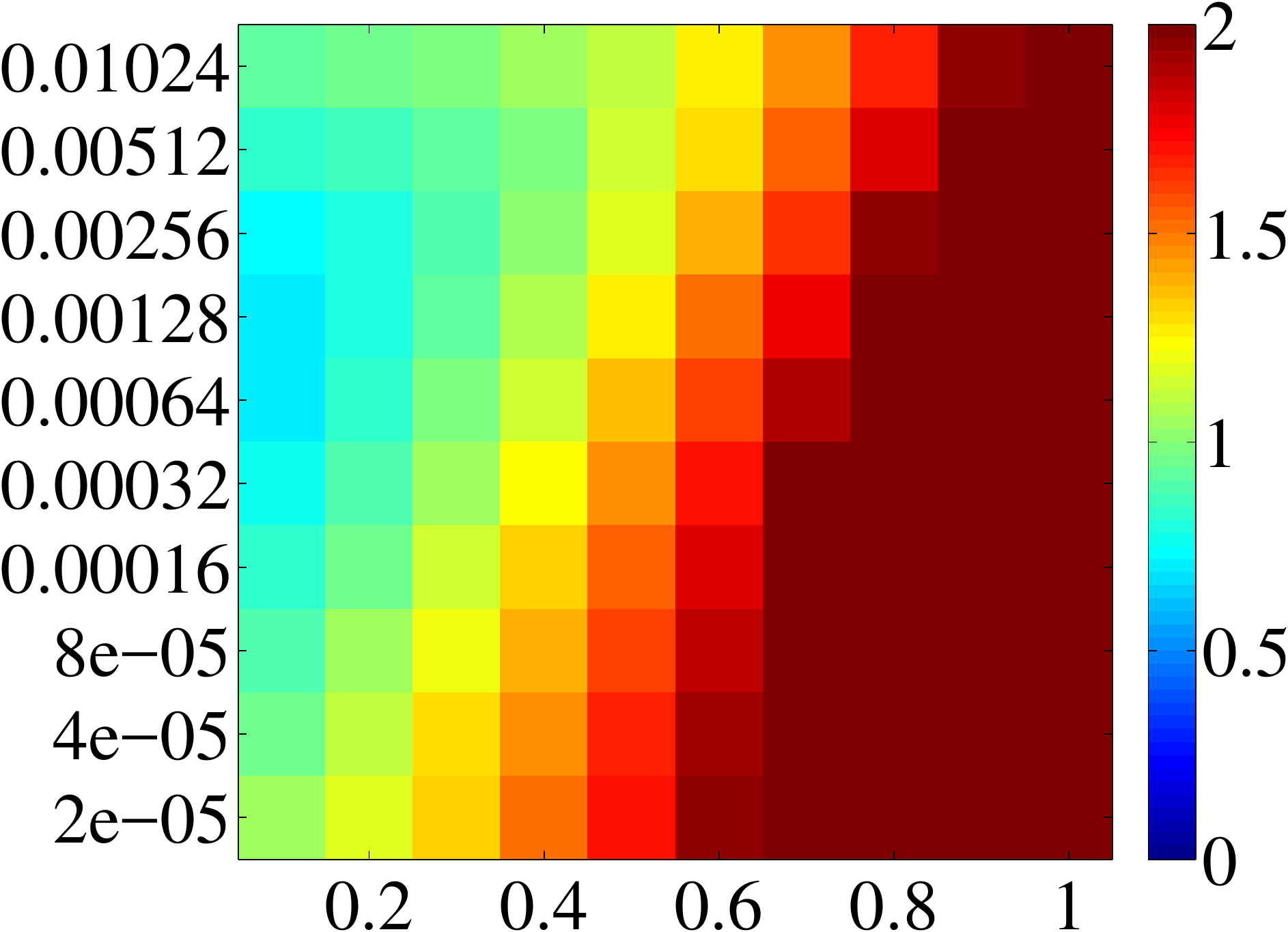} &
\includegraphics[width=0.43\linewidth]{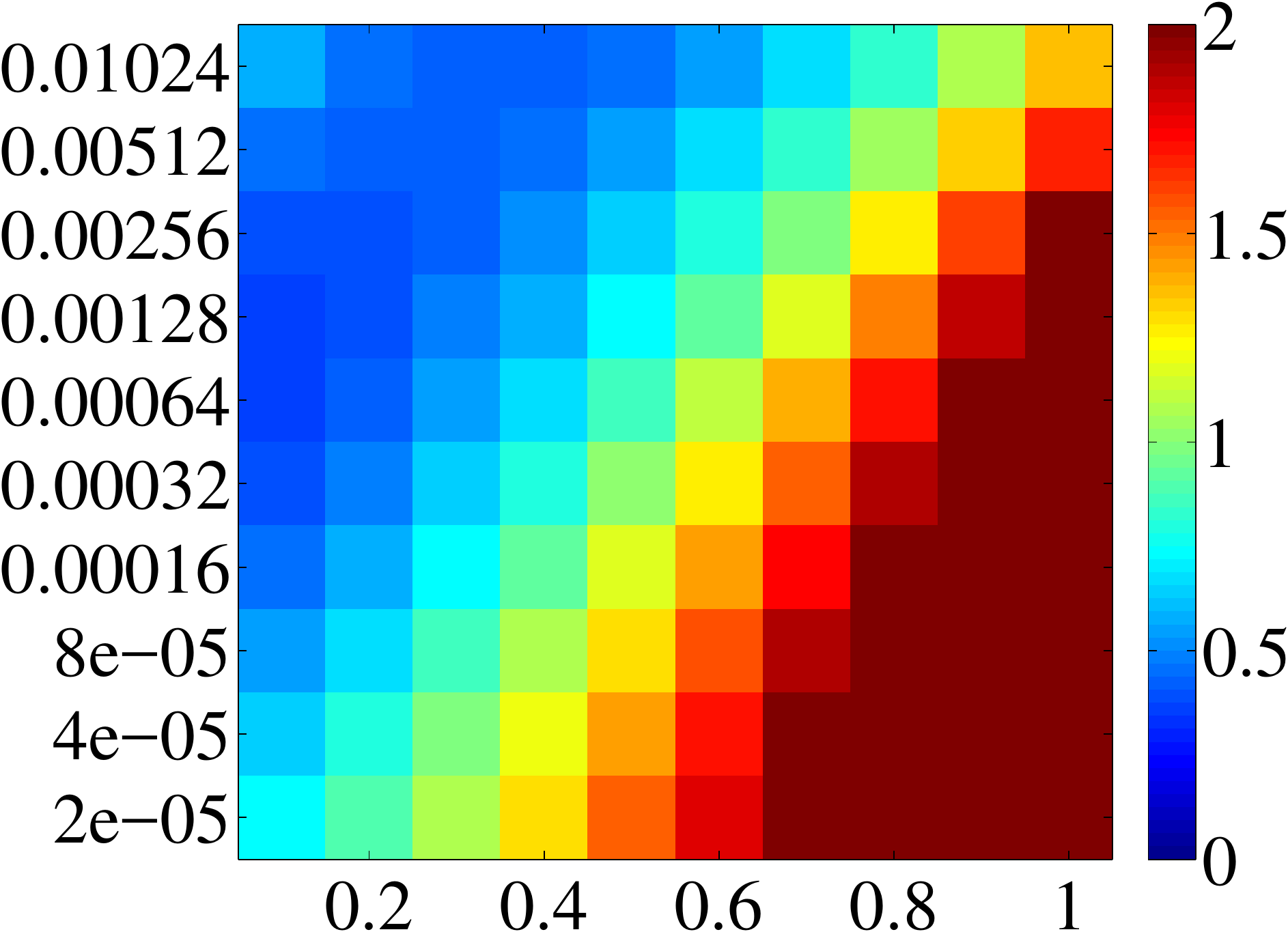} \\
(a) $f^{\textrm{IRLS}}(k_{gt}) / f^{\textrm{opt}}(k_{\delta})$ &
(b) $\rho_l(\hat{l}_{gt}^\textrm{IRLS}) / \rho_l(\hat{l}_{\delta}^{\textrm{opt}})$
\end{tabular}
}
\caption{The x- and y-axes of each plot represent $\alpha$ and $\lambda_l$, respectively. Values larger than 2 are clipped to 2 for better visualization.}
\label{fig:alpha_lambda_plots}
\vspace{-3.5mm}
\end{figure}

\Fig{alpha_lambda_plots} shows $f^\textrm{IRLS}(k_{gt}) / f^\textrm{opt}(k_\delta)$ and $\rho_l(\hat{l}_{gt}^\textrm{IRLS})/ \rho_l(\hat{l}_\delta^\textrm{opt})$ for different $\lambda_l$ and $\alpha$.
Note that the ratios $f^\textrm{IRLS}(k_{gt}) / f^\textrm{opt}(k_\delta)$ and $\rho_l(\hat{l}_{gt}^\textrm{IRLS})/ \rho_l(\hat{l}_\delta^\textrm{opt})$ present tighter bounds for $\alpha$ and $\lambda_l$ than the true bounds because $\hat{l}_{gt}^\textrm{IRLS}$ is a local optimum.
Despite these tighter bounds, \Fig{alpha_lambda_plots} shows that the ground truth blur kernel $k_{gt}$ is favored over $k_\delta$ by the energy function $f$ and the prior $\rho_l$ when $\alpha$ is small and $\lambda_l$ is large enough.
When $\lambda_l$ is too large, both $\hat{l}_{gt}$ and $\hat{l}_\delta$ become completely zero, so no longer distinguishable.

\begin{figure}[t]
\centering
\includegraphics[width=0.8\linewidth]{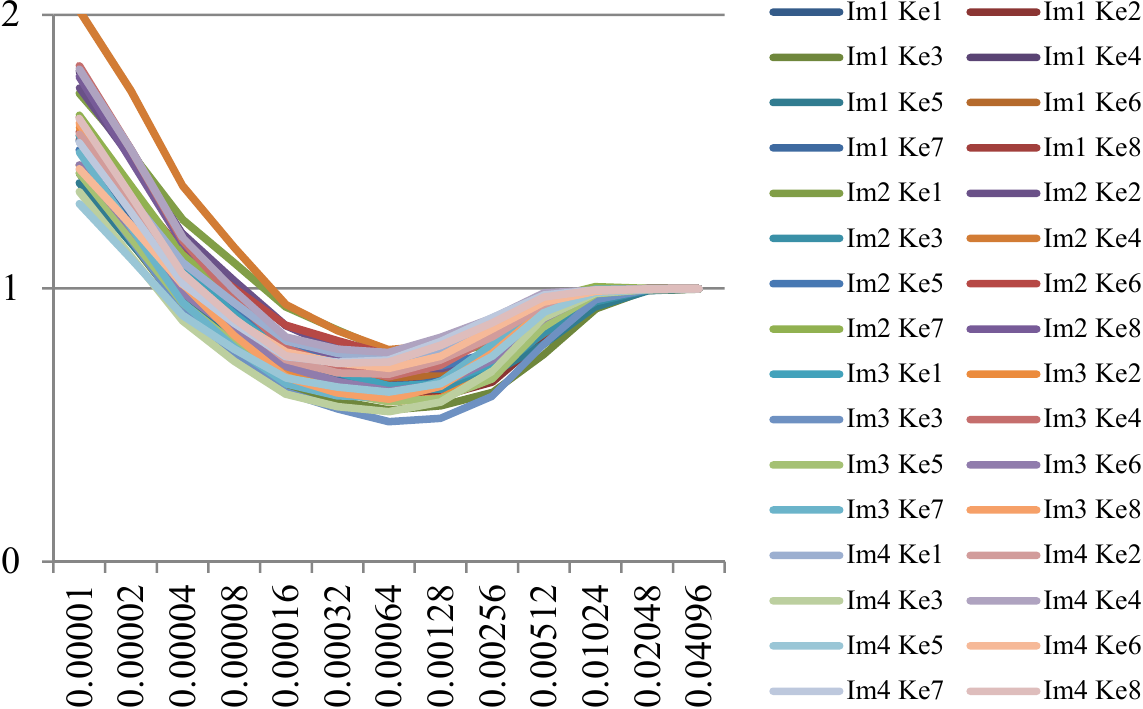}\\
\includegraphics[width=0.49\linewidth]{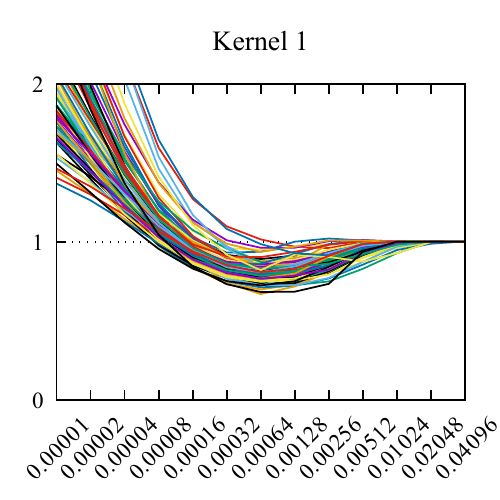}
\includegraphics[width=0.49\linewidth]{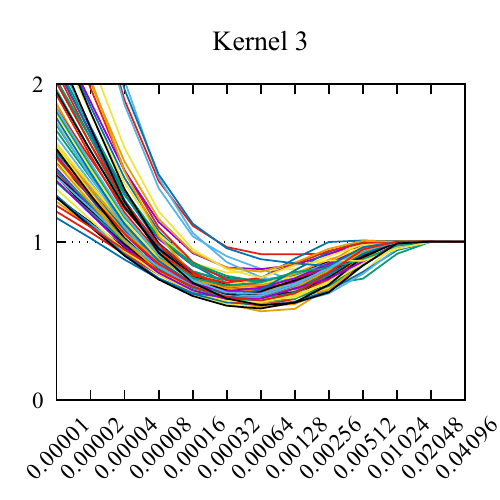}
\caption{$f^\textrm{IRLS}(k_{gt}) / f^\textrm{opt}(k_\delta)$ with respect to different $\lambda_l$'s. (Top: Levin et al.'s dataset~\cite{Levin-CVPR09}. Bottom: Sun et al.'s dataset~\cite{Sun-ICCP13}) $f^\textrm{IRLS}(k_{gt}) / f^\textrm{opt}(k_\delta)$ smaller than $1$ means that the ground truth blur kernel is preferred to the delta kernel by the energy function.}
\label{fig:gt_vs_delta}
\vspace{-2.5mm}
\end{figure}

To investigate the bounds for convergence more rigorously,
we compute the ratios $f^\textrm{IRLS}(k_{gt}) / f^\textrm{opt}(k_\delta)$ on two publicly avaiable datasets: Levin et al.'s~\cite{Levin-CVPR09} and Sun et al.'s~\cite{Sun-ICCP13}~(\Fig{gt_vs_delta}).
Levin et al.'s dataset consists of 32 real blurred images generated from four images and eight blur kernels.
On the other hand, Sun et al.'s consists of 640 synthetically blurred images generated from 80 sharp images ranging from natural scenes to man-made environments, and eight blur kernels.
In this experiment, we compute $f^\textrm{IRLS}(k_{gt}) / f^\textrm{opt}(k_\delta)$ for fixed $\alpha=0.1$ and different $\lambda_l$.
\Fig{gt_vs_delta} shows that the energy function favors the ground truth kernel over the no-blur solution for most images once $\alpha$ and $\lambda_l$ are properly set.\footnote{Refer to the supplementary material for the rest of the results.}
We can also observed that, while different blur kernels and images show different energy value ratios, they still show similar trends.
This indicates that a carefully chosen $\lambda_l$ can cover most of the images and the blur kernels.

\begin{figure}[t]
\centering
\includegraphics[width=0.8\linewidth]{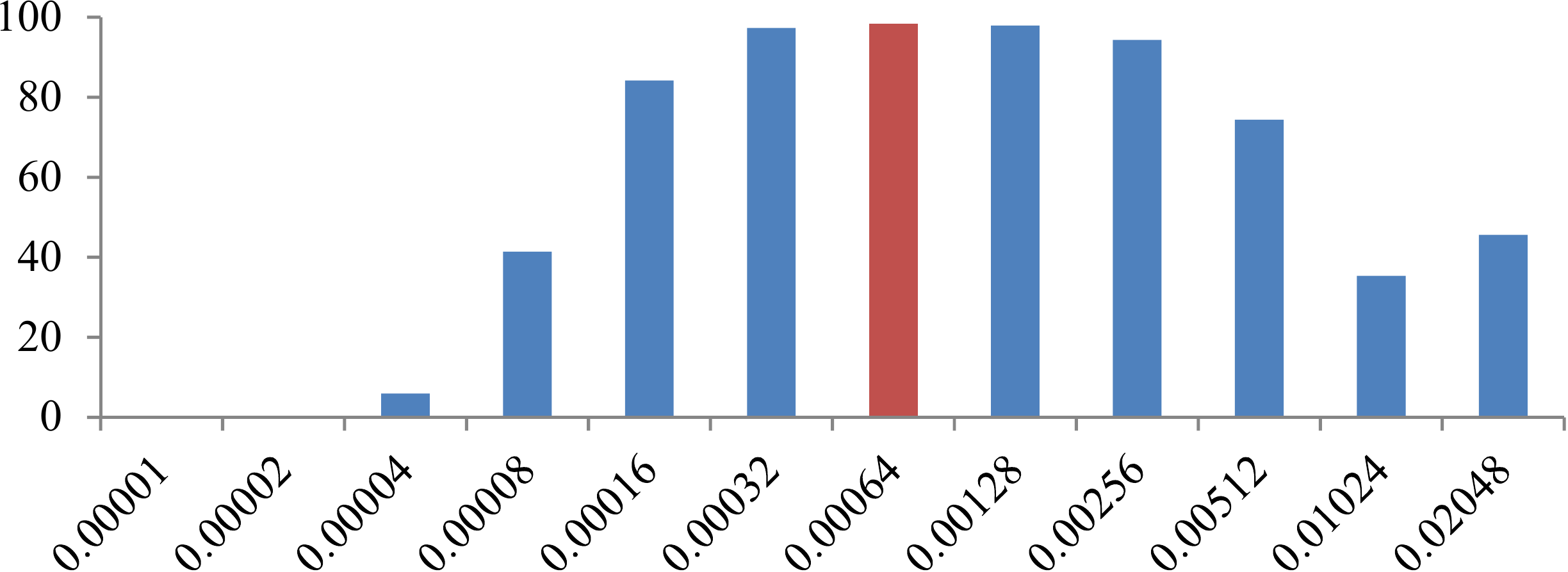}
\caption{Percentages of images in Sun et al.'s dataset~\cite{Sun-ICCP13} satisfying $f^\textrm{IRLS}(k_{gt}) / f^\textrm{opt}(k_\delta) < 1$ with different $\lambda_l$. $\lambda_l = 0.00064$ is the most effective to distinguish sharp and the no-blur solutions.}
\label{fig:energy_ratio_histogram}
\vspace{-2.5mm}
\end{figure}

It is also worth noting that some images have the ratio $f^\textrm{IRLS}(k_{gt}) / f^\textrm{opt}(k_\delta)$ above 1 for almost the entire range of $\lambda_l$, which indicates that the energy function is not able to distinguish the right solution and the no-blur one.
Such images have a relatively small number of edges, and previous methods often fail on such images.
Our results suggest that such failures cannot be avoided using different parameters, but instead a more improved algorithm is needed.
In the remainder of this paper, we consistently use $\alpha=0.1$ and $\lambda_l=0.00064$, which are shown to be the most effective to distinguish sharp and the no-blur solutions in these experiments, i.e., the largest number of images have $f^\textrm{IRLS}(k_{gt}) / f^\textrm{opt}(k_\delta) < 1$ under these parameters (\Fig{energy_ratio_histogram}).

\subsection{Global Optimum and Convergence Analysis}
\label{sec:global_optimum}

In \Sec{no_blur_condition}, we experimentally showed that a MAP based energy function can favor a sharp solution over the no-blur solution by comparing their energy values.
In this section, we investigate two questions: 1) does the true blur kernel actually correspond to the global optimum of \Eq{reformulated_energy_function}, and 2) how well does na\"{i}ve MAP based blind deconvolution perform compared to previous sophisticated methods?

Regarding the first question,
when $\lambda_l$ is set strong enough, a latent image obtained by the true blur kernel should have sharp edges and flat regions, minimizing $\rho_l(\hat{l})$ in \Eq{reformulated_energy_function}.
On the other hand, a different blur kernel usually causes blurry edges or ringing artifacts in its latent image, increasing $\rho_l$, and eventually its energy value.
It is hard to analytically prove this property because evaluation of \Eq{reformulated_energy_function} involves a complex non-linear optimization in \Eq{optimal_l_k}.
Instead, we provide a simple experiment with 1D blur kernels, and also experimentally show that minimizing \Eq{reformulated_energy_function} converges to the right solution.

Regarding the second question, previous successful methods adopt either explicit edge detection~\cite{Cho-SIGGRAPHAsia09,Xu-ECCV10,Sun-ICCP13}, edge reweighting~\cite{Shan-SIGGRAPH08}, changing parameters of the energy function~\cite{Shan-SIGGRAPH08,Wipf-JMLR14}, or variational Bayesian estimation~\cite{Fergus-SIGGRAPH06,Levin-CVPR09,Levin-CVPR11,Wipf-JMLR14}.
While such techniques may help improve their performances, we show that even a na\"{i}ve MAP approach can perform comparably despite lack of such components.

\begin{figure}[t]
\centering
\includegraphics[width=0.95\linewidth]{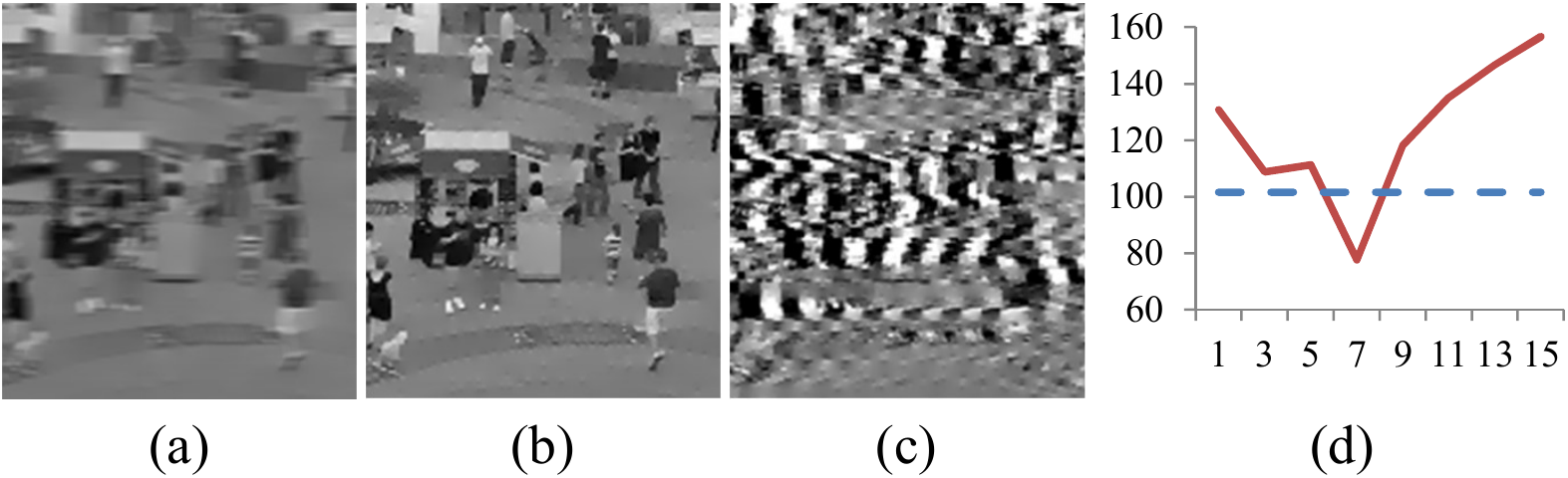}
\caption{(a), (b) and (c) show sparse latent image estimates $\hat{l}$ obtained using blur kernels of lengths 1, 7, and 15, respectively. The original blurry image is blurred by the blur kernel of length 7. (d) Solid red line: energy values $f^\textrm{IRLS}(k)$ of blur kernels of different lengths, and dashed blue line: $f^\textrm{opt}(k_\delta)$.}
\label{fig:global_optimum}
\vspace{-4.5mm}
\end{figure}

\Fig{global_optimum} shows a simple experiment to see whether the true blur kernel corresponds to the global optimum.
We first blur a sharp natural image using a 1D blur kernel of length $7$.
Then, we compute the energy values of blur kernels of different lengths.
\Fig{global_optimum}(d) shows the energy values of different blur kernels. The plot shows that the ground truth blur kernel is preferred by the energy function.

\begin{figure}[t]
\centering
\includegraphics[width=0.95\linewidth]{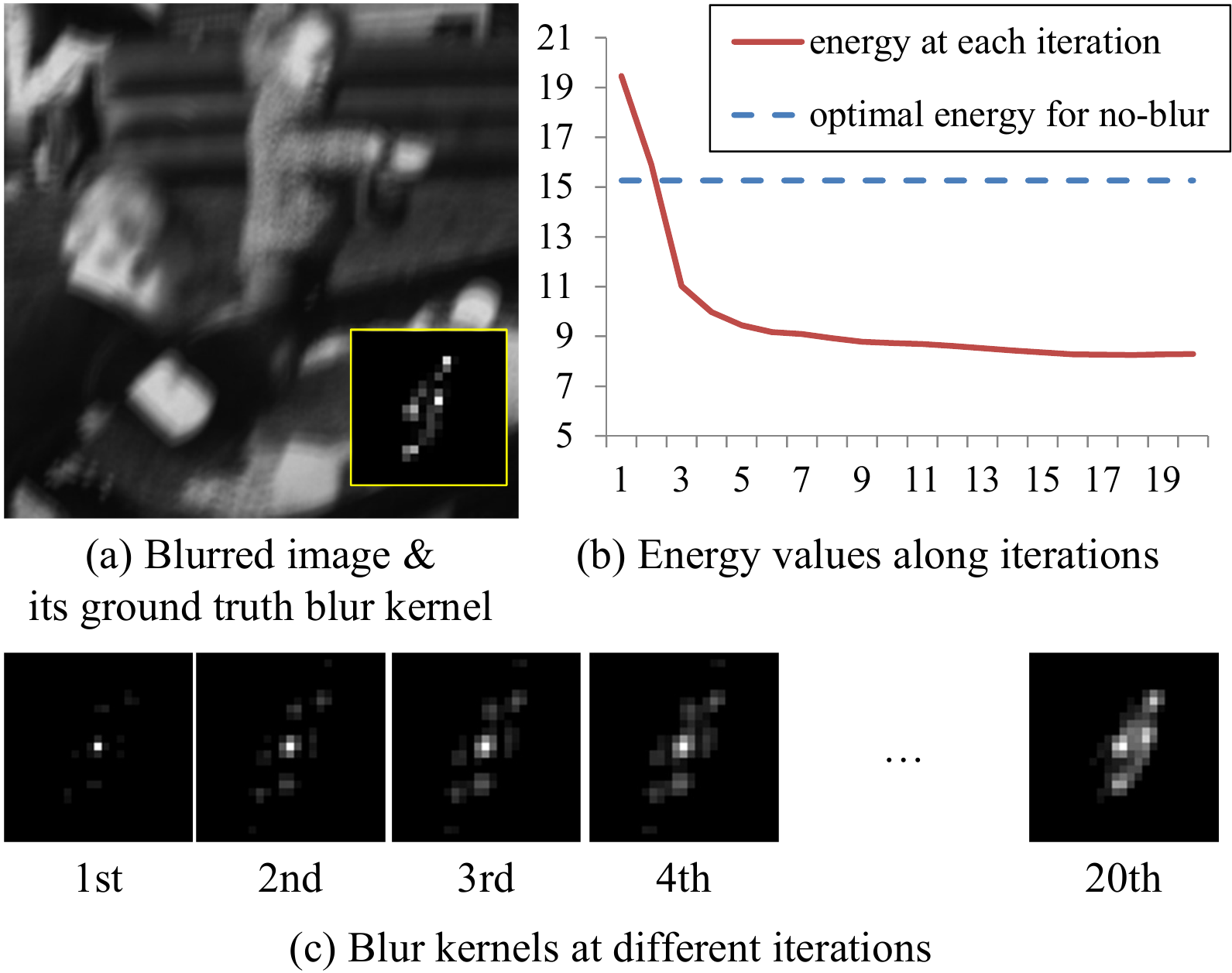}
\caption{Minimizing \Eq{reformulated_energy_function} converges to a sharp solution, which is close to the ground truth blur kernel.}
\label{fig:energy_minimization}
\vspace{-4.5mm}
\end{figure}

\begin{figure}[t]
\centering
\includegraphics[width=0.95\linewidth]{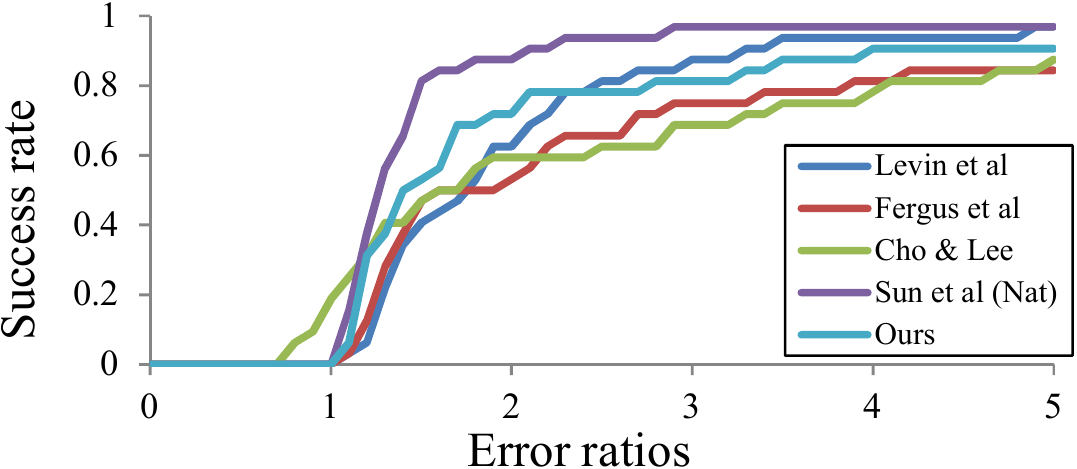}
\caption{Performance comparison with Levin et al.~\cite{Levin-CVPR09}, Fergus et al.~\cite{Fergus-SIGGRAPH06}, Cho \& Lee~\cite{Cho-SIGGRAPHAsia09}, and Sun et al.~\cite{Sun-ICCP13} using the cumulative error ratio histogram proposed by~\cite{Levin-CVPR09} and Levin et al.'s dataset~\cite{Levin-CVPR09}. Success rates of other methods are from \cite{Sun-ICCP13}.
}
\label{fig:full_blind_deconvolution}
\vspace{-3.5mm}
\end{figure}

Finally, we implement na\"{i}ve MAP based blind deconvolution, which optimizes \Eq{reformulated_energy_function}. Note that optimizing \Eq{reformulated_energy_function} is equivalent to optimizing \Eq{joint_energy_function} as:
\begin{eqnarray}
\min_{k,l} f(k,l) = \min_k \{\min_l f(k,l)\} = \min_k f(k).
\end{eqnarray}
Moreover, \Eq{reformulated_energy_function} yields the exactly same alternating optimization process described by \Eqs{l_step} and (\ref{eq:k_step}).
Given an estimate of $k$, we compute $\hat{l}_k$ by optimizing \Eq{l_step}, and then update $k$ by optimizing \Eq{reformulated_energy_function}, which is equivalent to optimizing \Eq{k_step}.
We implemented single- and multi-scale versions, and set $\lambda_k = 0.001$.
\Fig{energy_minimization} shows that the single-scale version can converge to a solution close to the true kernel whose energy is lower than that of the no-blur solution.
We conducted performance comparison of the multi-scale version using Levin et al.'s dataset~\cite{Levin-CVPR09} (\Fig{full_blind_deconvolution}).
Although our result is poorer than \cite{Sun-ICCP13}, which is based on patch-based priors, it is still comparable to the others.
This shows that even a na\"{i}ve MAP approach can perform comparably to the other sophisticated methods.
Furthermore, while converging to the true kernel does not necessarily mean that the true kernel is the global optimum, it indicates that the true kernel is preferred to other kernels estimated through the optimization process.

%

\section{Energy Function as a Kernel Quality Metric}
\label{sec:other_applications}

Besides estimation of blur caused by camera shakes, there are many problems related to image blur, such as defocus estimation~\cite{Zhuo-PR11}, lens blur estimation~\cite{Schuler-ECCV2012}, blur kernel size detection~\cite{Liu-VisComp15}, fusion of deblurring results obtained by different blur kernels~\cite{Liu-SIGGRAPHAsia13}, etc.
In those problems, it is essential to have a metric for evaluating the quality of a blur kernel.
Unfortunately, because there has been no universal metric proven to work, solutions for different problems defined their own metrics.

The energy function in \Eq{reformulated_energy_function} is a function of a blur kernel, which properly gives a lower energy to a better blur kernel when $\lambda_l$ is properly set.
Therefore it provides a simple and effective metric to compare blur kernels, which can be applied universally to different problems.
While the idea of using an energy function as a metric may sound straightforward and obvious, this simple idea was not possible because of mainly two reasons.
First, the original joint energy function in \Eq{joint_energy_function} involves two variables $l$ and $k$, so it was rather unclear how to utilize the energy function to other problems.
Second, it was unclear whether and when the energy function in \Eq{joint_energy_function} favors the sharp solution over the no-blur one.
Our modification to the energy function and analysis in \Sec{convergence} resolve these two issues and make the above idea possible.
In this section, we present three examples as possible applications of the energy function.

\subsection{Automatic Blur Kernel Size Selection}

Most blind deconvolution methods require the size of a blur kernel as input.
An input kernel size smaller than the actual blur size results in erroneous kernel estimation.
On the other hand, a too large kernel size increases the degree of freedom of kernel estimation, which may lead to an unstable and erroneous result.
However, it is not an easy task for a user to select a proper size.
There have been a few attempts to automatically find a proper kernel size~\cite{Liu-SIGGRAPHAsia13,Liu-VisComp15}.
Liu et al.~\cite{Liu-SIGGRAPHAsia13} deblurred an image with a set of different blur kernels of different sizes and found a proper kernel size using their deblurring quality metric trained from crowd-sourced user study data.
Recently, Liu et al.~\cite{Liu-VisComp15} proposed a kernel size estimation method, which estimates a kernel size from the autocorrelation of the edge map of a blurred image.

\begin{figure}[t]
\centering
\includegraphics[width=1\linewidth]{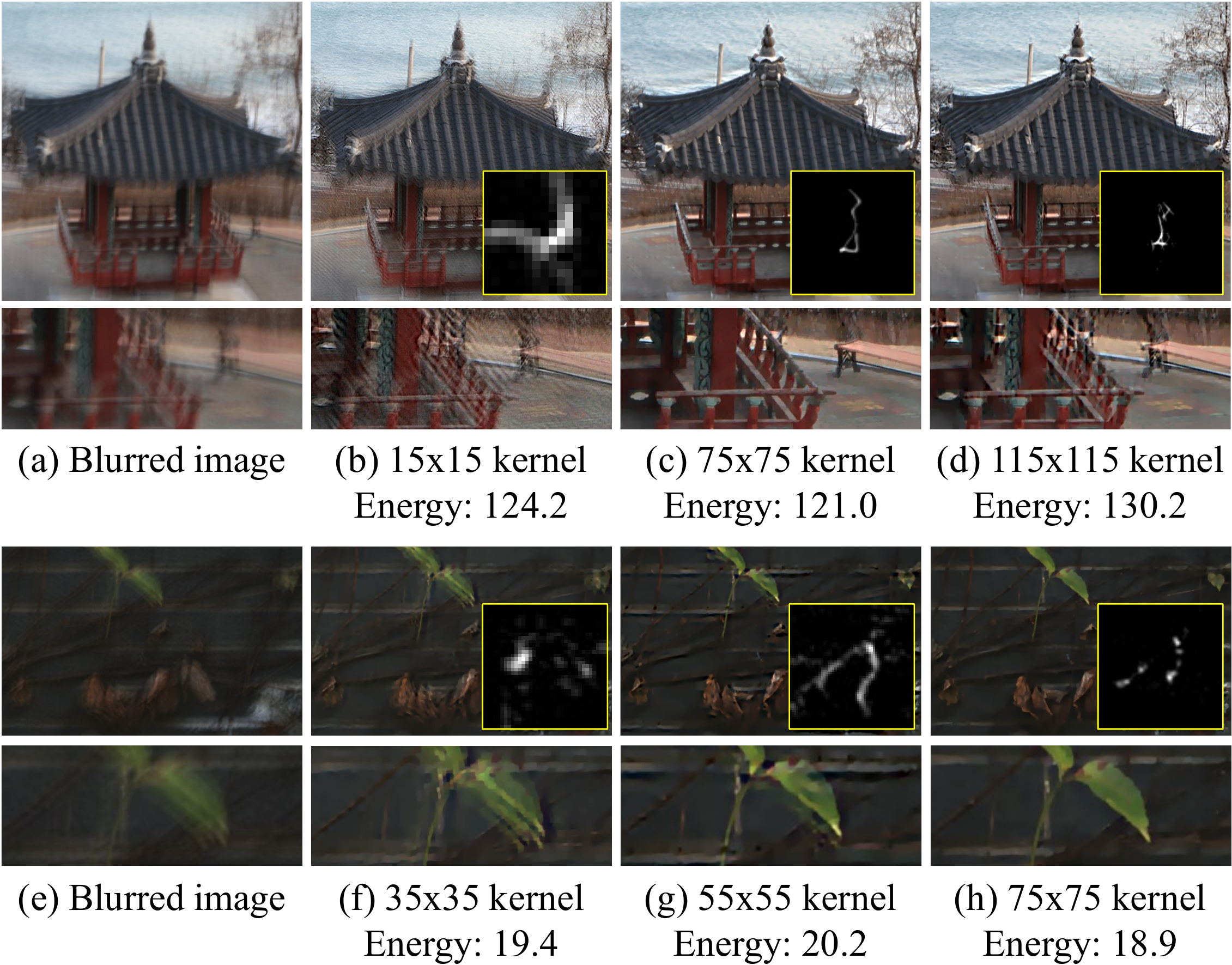}
\caption{(a) \& (e) Blurred images. (b)-(d) \& (f)-(h) Estimated blur kernels of different sizes and their corresponding latent images. All the deblurring results are obtained using \cite{Cho-SIGGRAPHAsia09}, and their energy values are computed using \Eq{reformulated_energy_function}.
The sizes of the blur kernels in (b), (f), and (g) are too small, so incorrect kernels are estimated. On the other hand, a too large kernel size in (d) also results in incorrect kernel estimation. The energy function in \Eq{reformulated_energy_function} can properly distinguish the correct solutions (c) and (h) from the others.}
\label{fig:blur_kernel_size}
\vspace{-3.5mm}
\end{figure}

The energy function in \Eq{reformulated_energy_function} provides a simpler way to find out a proper kernel size.
Similarly to \cite{Liu-SIGGRAPHAsia13}, we first estimate blur kernels of different sizes.
Then, we compute their energies and choose the kernel with the smallest energy.
\Fig{blur_kernel_size} shows an example.

Recall that this simple approach for comparing different kernels has been made possible due to our analysis.
Our reformulated energy function states that a properly estimated latent image must be used for computing the energy instead of any arbitrary latent image, e.g., a naturally-looking latent image obtained from a previous method.
We also showed that parameters must be properly set in order to make the energy function favor the right solution.
For example, an inappropriate $\lambda_l=0.00001$ produces energy values 13.0, 14.4, and 16.5 for the kernels in \Fig{blur_kernel_size}(b), (c), and (d), respectively, and causes the energy function to prefer the smallest kernel, which is close to the no-blur solution.

\subsection{Blur Kernel Estimation from Light Streaks}

\begin{figure}[t]
\centering
\includegraphics[width=0.95\linewidth]{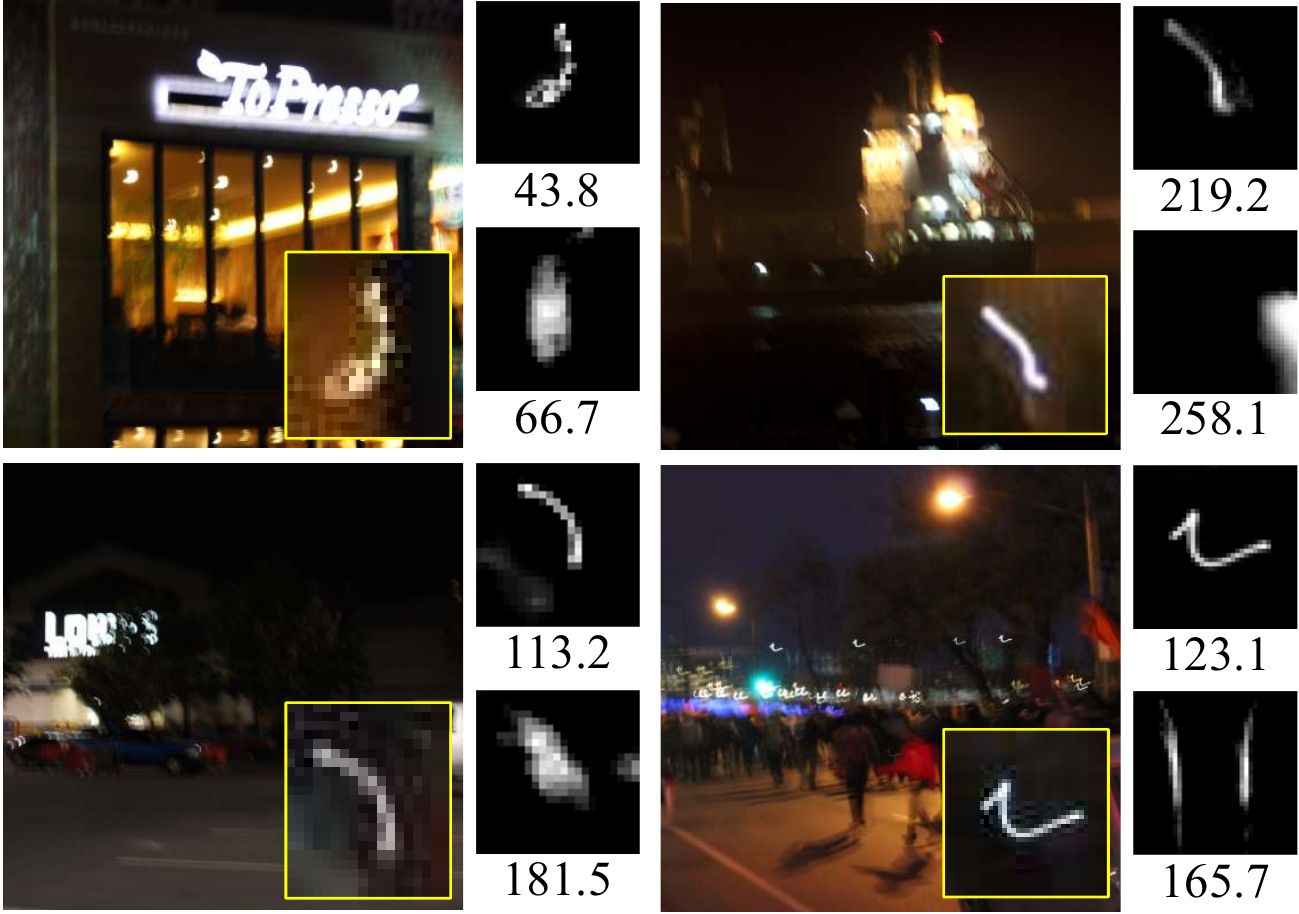}
\caption{For each image, left: Blurred images with light streaks, and a magnified patch of a light streak, which reflects the shapes of blur kernels. Right: The best and worst light streak patches selected by \Eq{reformulated_energy_function} and their corresponding energy values.}
\label{fig:light_streaks}
\vspace{-2.5mm}
\end{figure}

Images blurred by camera shakes often have light streaks, which are caused by blurred light bulbs, flash lights, reflected light, etc~(blurred images in \Fig{light_streaks}).
Such light streaks provide useful information about the shape of the blur kernel, and a couple of methods have been proposed to use light streaks for blur kernel estimation.
Hua and Low~\cite{Hua-ICIP11} proposed an interactive method, where the user manually draws a small bounding box for a light streak, and then, the system extracts a blur kernel using heuristic image processing operations.
Zhe et al.~\cite{Zhe-CVPR14} presented a more sophisticated method.
Their method automatically detects light streaks from a blurred image, then uses the detected light streaks to estimate a blur kernel.
In order to detect light streaks, their method first uses a set of heuristic rules for detecting light streak patches.
Then the best light streak patch is selected based on the power-law of natural images,
and used for detecting additional light streak patches.

Instead of the power law, which is known to be sensitive to strong edges~\cite{Yue-ECCV14,Liu-VisComp15}, \Eq{reformulated_energy_function} provides a more direct measure to select the best light streak patch.
Similarly to \cite{Zhe-CVPR14}, we first find a set of candidate light streak patches using heuristic rules.
In our experiment, we use the code of the authors of \cite{Zhe-CVPR14} to find an initial candidate set.
Then, instead of the power law based metric, we compute their energy values using \Eq{reformulated_energy_function}, and choose the one with the lowest energy.
\Fig{light_streaks} shows an example.
For each blurred image in \Fig{light_streaks}, we show the best and worst patches according to their energy values.
While the best patches selected by \Eq{reformulated_energy_function} include proper light streak patches reflecting blur kernels,
the worst patches are far from the true kernels.


\subsection{Defocus Estimation}

Defocus blur is caused by shallow depth-of-field of an imaging system, and it is often spatially varying.
As the amount of defocus blur is related to the distance from the camera to the target object, defocus information can be useful for depth estimation, salient region estimation, foreground/background segmentation, digital refocusing, etc.

However, estimating a defocus map from a single image is a challenging task, as the amount of defocus blur can be different at each pixel.
To overcome such difficulty, previous methods proposed several different features to detect the amount of blur.
Tai and Brown~\cite{Tai-ICIP09} proposed a measure based on a local contrast prior, which utilizes the relationship between local image contrast and image gradients.
Zhuo and Sim~\cite{Zhuo-PR11} re-blur the input defocused image with a Gaussian blur kernel, and use the ratio between the gradients of the input and the re-blurred images to estimate a defocus map.

\Eq{reformulated_energy_function} can also be used for estimating the amount of defocus blur.
We first assume that the shape of defocus blur is already known, but its size is unknown and spatially varying, e.g., spatially-variant disk filters.
As \Eq{reformulated_energy_function} is based on a sparsity prior, we can compare different blur kernels more robustly on strong edges.
Thus, we first detect edges using Canny edge detector, and compare energy values on the detected edge pixels.
The energy value of a blur kernel on an edge pixel is defined as the energy value on a local image region centered at the edge pixel.
As a result, we obtain a sparse defocus map, where defocus blur sizes are estimated only on edge pixels.
We then spatially propagate this defocus information to other pixels using the matting Laplacian algorithm~\cite{Levin-PAMI08}, as done in \cite{Zhuo-PR11}.

\begin{figure}[t]
\centering
\includegraphics[width=0.95\linewidth]{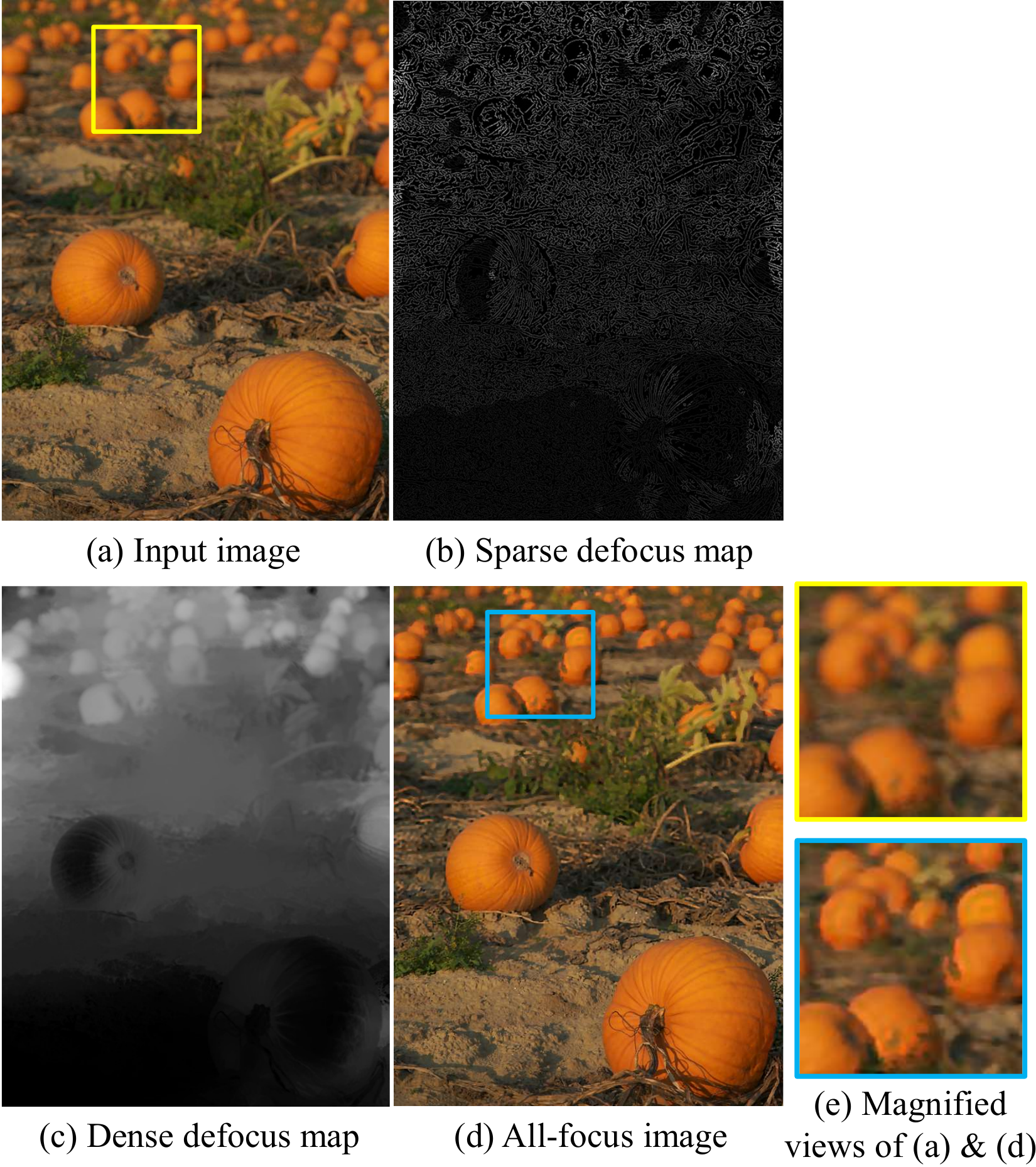}
\caption{Real defocus example.}
\label{fig:defocus_blur}
\vspace{-2.5mm}
\end{figure}

\begin{figure}[t]
\centering
\includegraphics[width=1\linewidth]{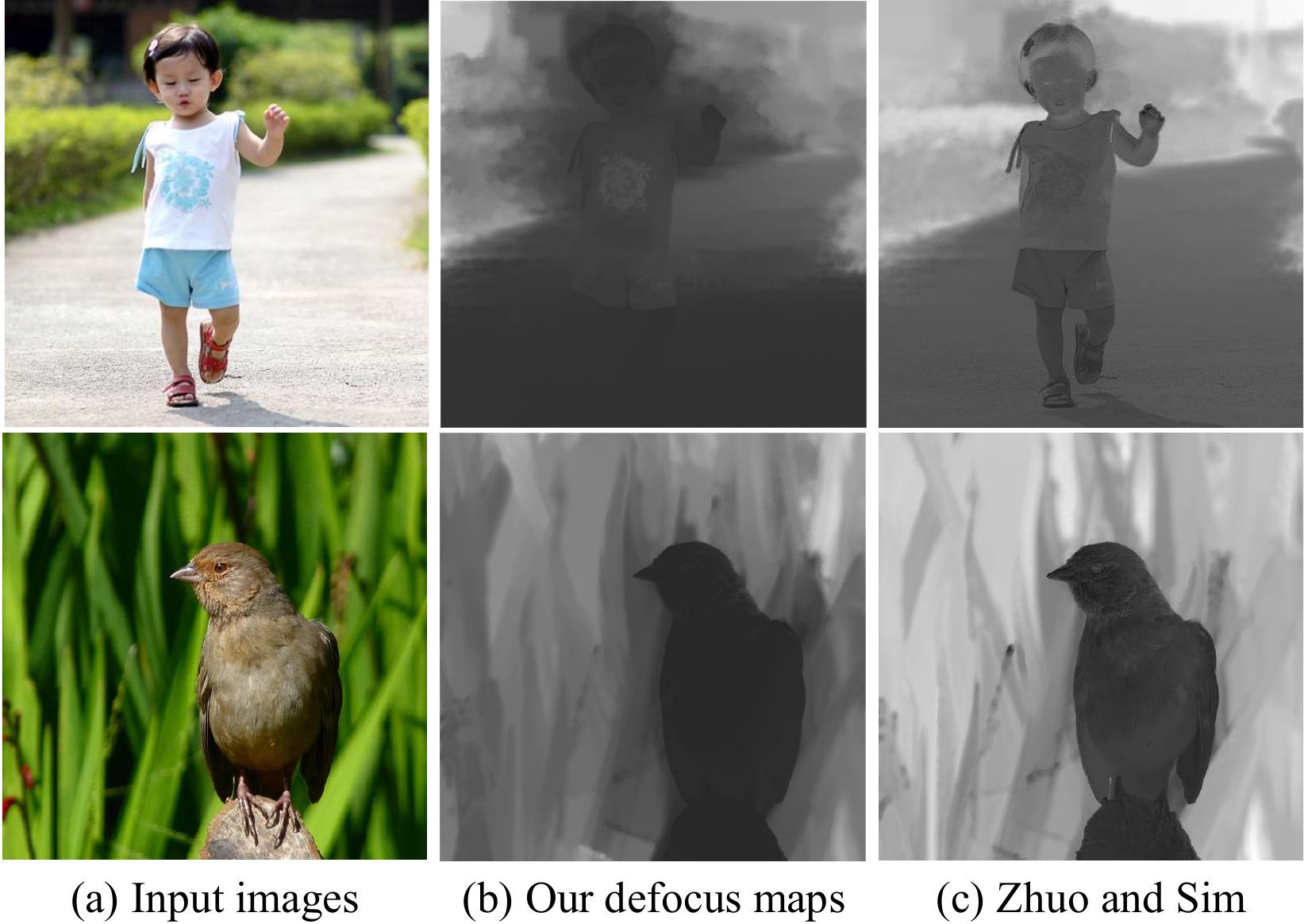}
\caption{Additional real defocus examples. While \Eq{reformulated_energy_function} is a universal metric, which is not designed for defocus estimation, it produces comparable results to Zhuo and Sim~\cite{Zhuo-PR11}.}
\label{fig:defocus_blur_other_ex}
\vspace{-2.5mm}
\end{figure}

\Fig{defocus_blur} shows a defocus estimation example.
\Fig{defocus_blur}(b) is a sparse defocus map estimated from \Fig{defocus_blur}(a). Brighter pixel means larger defocus blur.
\Fig{defocus_blur}(c) shows a full defocus map obtained from \Fig{defocus_blur}(b) using the matting Laplacian algorithm.
As the upper part of the image is more distant and more defocused, the estimated defocus map shows brighter pixels in that part.
\Fig{defocus_blur}(d) is an all-focused result obtained using the defocus map in \Fig{defocus_blur}(c).
\Fig{defocus_blur_other_ex} shows additional examples.
While \Eq{reformulated_energy_function} is a universal metric, which is not specially designed for defocus estimation, it produces comparable defocus maps to Zhuo and Sim's method~\cite{Zhuo-PR11}.

\section{Conclusions}

In this paper, we analyzed the convergence of MAP based blind deconvolution, and showed that the energy function is the key to the success of previous MAP based approaches.
To this end, we introduced a reformulated energy function. Then, we analyzed conditions for avoiding no-blur solution, and showed that the energy function can converge to the right solution.
We also demonstrated that the reformulated energy function can be used as a simple and effective metric to compare different blur kernels with three examples.
In our experiments, we used IRLS for solving \Eq{optimal_l_k}, which requires some amount of computation.
One interesting future work would be to develop an efficient latent image estimation method for solving \Eq{optimal_l_k}
while guaranteeing \Eqs{condition1_for_avoiding_no_blur} and (\ref{eq:condition2_for_avoiding_no_blur}).

\vspace{-1em}
{\small
\paragraph{Acknowledgements} This work was supported by the DGIST Start-up Fund Program of the Ministry of Science, ICT and Future Planning(2017040005).
It was also supported by the Ministry of Science and ICT, Korea, through IITP grant (R0126-17-1078) and NRF grant (NRF-2014R1A2A1A11052779).
}
{\small
\bibliographystyle{ieee}
\bibliography{main}
}

\end{document}